\documentclass[conference]{IEEEtran}
\IEEEoverridecommandlockouts
\usepackage{cite}
\usepackage{amsmath,amssymb,amsfonts}
\usepackage{algorithmic}
\usepackage{graphicx}
\usepackage{textcomp}
\usepackage{xcolor}
\usepackage{multirow}
\usepackage[
	breaklinks=true,		
	hypertexnames=false 
]{hyperref}

\def\BibTeX{{\rm B\kern-.05em{\sc i\kern-.025em b}\kern-.08em
    T\kern-.1667em\lower.7ex\hbox{E}\kern-.125emX}}
\begin{document}

\title{Towards Varroa destructor mite detection using a narrow spectra illumination}

\author{\IEEEauthorblockN{1\textsuperscript{st} Samuel Bielik}
\IEEEauthorblockA{\textit{Department of Control and Instrumentation} \\
\textit{Brno University of Technology}\\
Brno, Czech Republic \\
Samuel.Bielik@vut.cz}
\and
\IEEEauthorblockN{2\textsuperscript{nd} Šimon Bilík}
\IEEEauthorblockA{\textit{Institute for Research and Applications of Fuzzy Modeling} \\
\textit{University of Ostrava}\\
Ostrava, Czech Republic \\
simon.bilik@osu.cz}
}

\maketitle

\begin{abstract}
This paper focuses on the development and modification of a beehive monitoring device and \textit{Varroa destructor} detection on the bees with the help of hyperspectral imagery while utilizing a U-net, semantic segmentation architecture, and conventional computer vision methods. The main objectives were to collect a dataset of bees and mites, and propose the computer vision model which can achieve the detection between bees and mites.

\end{abstract}


\section{Introduction}
Bees are one of the most important creatures in the world, so humanity has to protect them. One of the bee diseases is varroosis, which is caused by \textit{Varroa destructor}. Overpopulation of this mite in beehive can ruin a whole beehive. Mechanical monitoring of the beehives is time-consuming. \textit{Varroa destructor} mites are much smaller than bees and they have similar color with the bees, so it can be very hard to detect this mite on body of the bee using common visible light. This paper describes a new approach to \textit{Varroa destructor} monitoring in the beehives with utilizing illumination with particular wavelengths. Our goal is to detect mites on bees which are flying into the beehives.

\section{Related Works}
Kim Bjerge et. al.~\cite{articleBjerge} developed device for automatic beehive inspection, which consists of camera and illumination unit. The device is mounted on the entrance of the beehive, where the bees pass through the parallel tunnels, which constraint their velocity and direction of movement. The research team also analysed 19 different wavelengths of visible and near infrared light in the range between 375 - 970 nm. The best wavelengths for resolution between bees and Varroa mites were considered as 450, 570 and 780 nm. The images are processed outside of the device and are not processed in real time. The algorithm first finds bees while using the Implicit shape model, and afterwards, it tries to find a Varroa mites on the segmented bees only using the convolutional neural network. The proposed solution also allows counting of the bees. 

In article~\cite{articleDumaHyperspectral}, Zina-Sabrina Duma et. al. aimed to determine the best wavelengths to distinguish between bees and mites from hyperspectral data captured using the Specim IQ camera. The bees and mites were illuminated by a custom multispectral LED unit, where brightness of individual LED wavelengths was adapted to Specim IQ hyperspectral camera chip sensitivity. To find the best wavelengths, the authors utilized the Principal component analysis (PCA) with K-means++ followed by the Kernel flow partial least squares. The best wavelengths were considered as 493, 499, 508 and 797 nm. 

U-net is a semantic segmentation architecture proposed by Olaf Ronneberger et. al. ~\cite{UNetPaper}. Semantic segmentation is a computer vision task in which the model assigns every pixel to a particular category. In U-net architecture, every cell in decoder is combined with corresponding cell from encoder. The predictions of the model are more exact thanks to this. One cell consists of two convolutional layers and max pooling layer in encoder or upsampling convolution in decoder. Every convolution si followed by ReLu activation function. This architecture does not have fully connected layer.

\section{Materials and methods}

\subsection{Experimental hardware}
\begin{figure}[!htb]
    \centering
    \includegraphics[width = 0.5\textwidth]{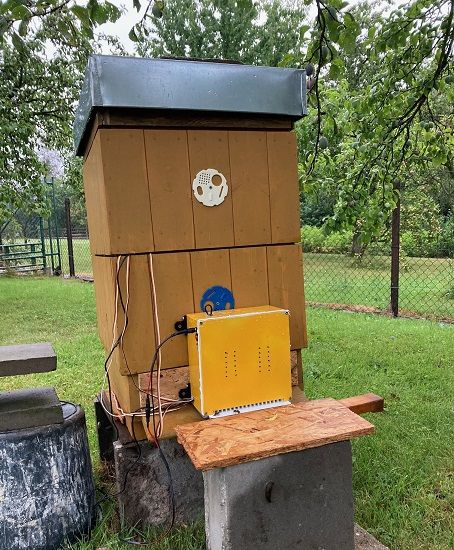}
    \caption{Fčielka-Thor 3000 mounted on the beehive.}
    \label{fig:Vcielkator}
\end{figure}

To collect a dataset, we modified an existing beehive monitoring device Fčielka-Thor 2000, which is described in~\cite{VcelkatorNevlacil}. The current version is Fčielka-Thor 3000 (Figure \ref{fig:Vcielkator}).

\begin{figure}[!ht]
    \centering
    \includegraphics[width = 0.5\textwidth]{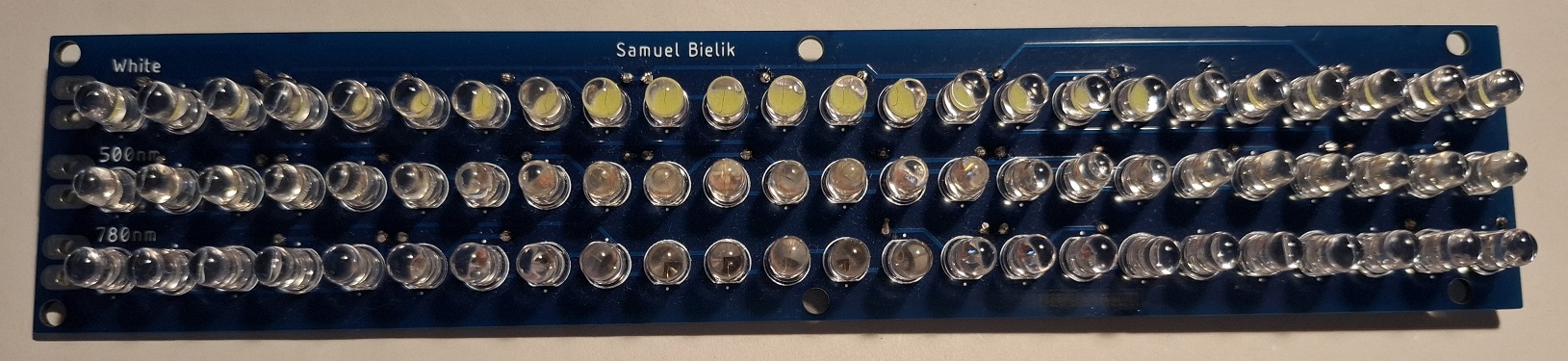}
    \caption{LED illumination unit.}
    \label{fig:Ill_Unit}
\end{figure}

\begin{figure}[!htb]
    \centering
    \includegraphics[width = 0.5\textwidth]{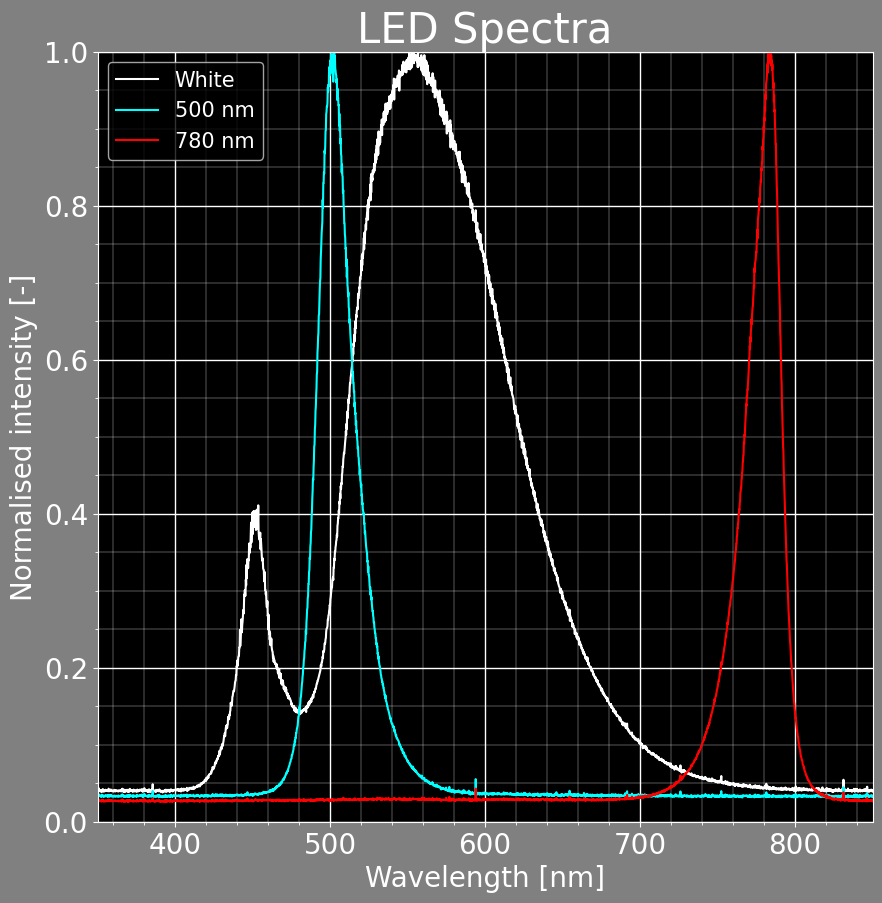}
    \caption{Measured LED spectra.}
    \label{fig:Spectra}
\end{figure}

We modified the illumination unit (Figure \ref{fig:Ill_Unit}), which now comprises three rows, each containing 24 LEDs emitting at 500 nm (turquoise), 780 nm (infrared) as proposed in~\cite{articleDumaHyperspectral}, and cold white for the general purposes. This unit is controlled by three PWM modules consisting of two parallelly connected MOSFET transistors AOD4184A, Raspberry Pi Pico and Raspberry Pi 4B. We measure the emitted light spectra of the utilized LEDs, which is shown in the Figure~\ref{fig:Spectra}. Our goal was to propose serial-parallel connection, in which the LED voltage is 0.1 V lower than declared LED maximum in datasheet. 

Photos were captured by Raspberry Pi HQ camera with the removed IR filter from the sensor. In the previous version, photos were captured continuously, and only image frames different from the last one were saved on the disk. In this version, the device waits for the button-triggering signal, and on each trigger, it takes three photos of three different illumination (white, infrared and turquoise). The photos are processed with the image calibrator module after the capture. The device also contains microphone and environment sensors (such as temperature, humidity and $CO_2$ ), but these data were not used in the current research. 

In the expected use-case, the bees will walk through the tunnels, which will separate them for an easier processing. Bees will be captured from the above in three different images using one selected illumination.

\subsection{Dataset}
Our dataset consists of 647 photos and is divided into 2 main parts: Photos of dead bees and dead mites before treatment and after treatment with fumigation. Both parts have 3 categories: Bees, mites and bees with mites. Bee and mite samples were collected in November 2024 and during the dataset capture, they were approximately 2-3 weeks old.

\begin{figure}[!htb]
    \centering
    \includegraphics[width = 0.5\textwidth]{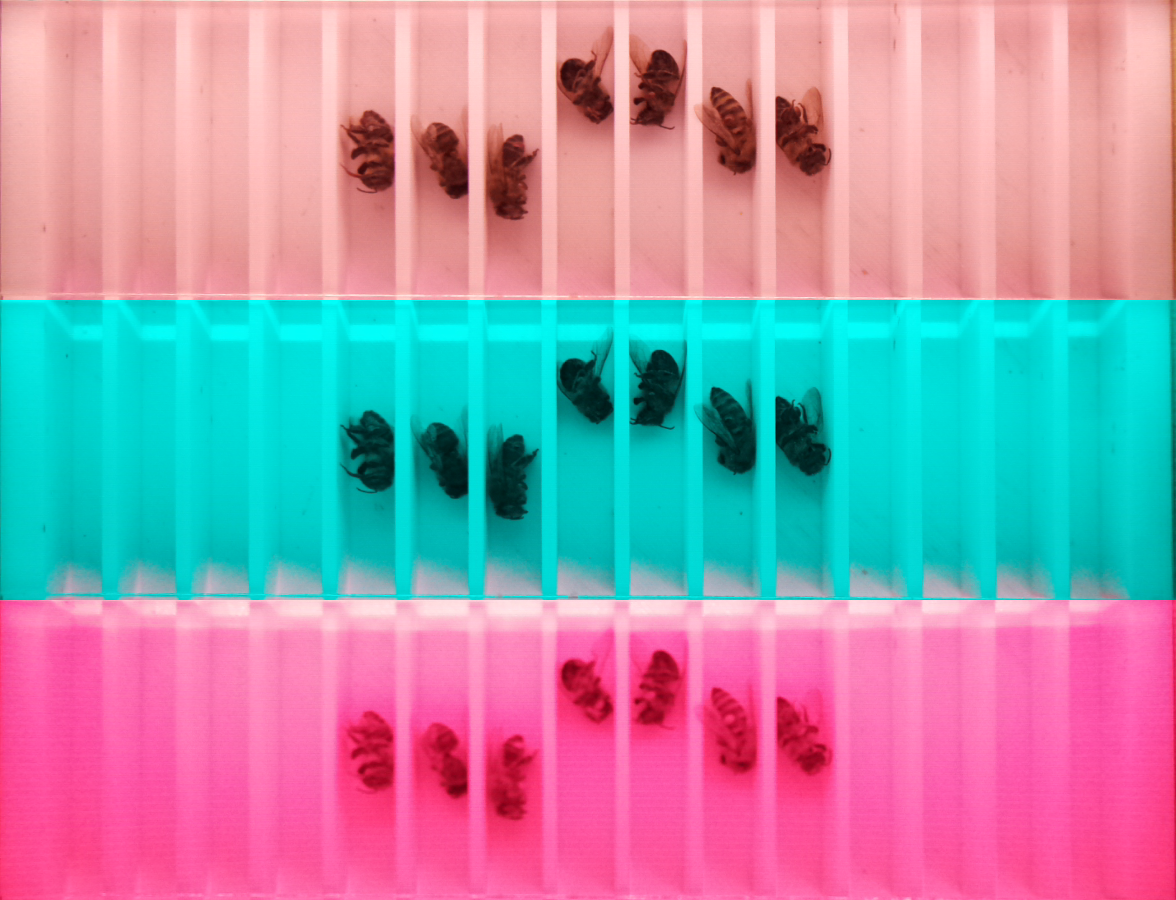}
    \caption{Example of photos in dataset.}
    \label{fig:DatasetExample}
\end{figure}

All samples were collected in the location of Těšínky, CZ. More details about the dataset are shown in Table~\ref{tab:Pocty}. To collect the dataset, we used around 25 bees and 20 Varroa mites in total. The photos of bees were taken from ventral, dorsal, left and right side. An example of all illumination of our dataset is shown in Figure~\ref{fig:DatasetExample}. All images have a uniform size of $1116 \times 300$ pixels.

The dataset can be found in~\cite{BeeDSHS2}. To utilize the dataset for the U-net training, we annotated it in LabelStudio tool~\cite{LabelStudio}. Export format was set as the three channels image, where every channel belongs to another category (blue - background, green - bee and red - mite). Because the bees were dead, we could use the same masks for the each of the illumination colors. Images were also annotated in the YOLO dataset format using the bees and mites classes. 

\begin{table}[!h]
    \caption{Number of photos in each category}
    \begin{center}
        \small
        \begin{tabular}{|c|c|c|c|}
            \hline
             Category&  Mite&  Bee& Bee with mite\\
             \hline\hline
             Before treatment& 78& 110 & 113\\
             \hline
             After treatment& 113& 113 & 120\\
             \hline
    \end{tabular}
    \label{tab:Pocty}
    \end{center}
\end{table}

\subsection{Evaluation metrics}

To evaluate the U-Net segmentation masks and the mite detections, we designed our own metric called Satisfied Bee Metric (SBM). As we just need to know, if the mite is present on the bee and we do not have to know its exact position, use of the common metrics could be misleading.

For every mite in prediction, the algorithm looks for intersection with the mite in the corresponding ground truth image. If a match occurs, the detection is considered as a true positive case. If a no intersection is found, it is considered as a false positive case. Afterwards, the algorithm takes every mite in ground truth image and searches intersection in the predicted image. If an intersection is found, nothing happens as the mite was counted in previous search, but if not, this case is considered as a false negative.

\section{Experiment description}

\subsection{Conventional computer vision based approach}

\label{Se:ConvApp}

As the mites under turquoise illumination almost coincide with bees, we also tried conventional methods without an use of a neural network. Conventional methods have an advantage in their speed and explainability but they also have a great disadvantage - as the live bees will move, such measurements and experiments will be challenging. These problems will result in different positions of bees in infrared and turquoise image. As we do not have a dataset on live bees, the experiments were performed only with static bees.

\begin{figure*}[!t]
    \centering
    \includegraphics[width = \textwidth]{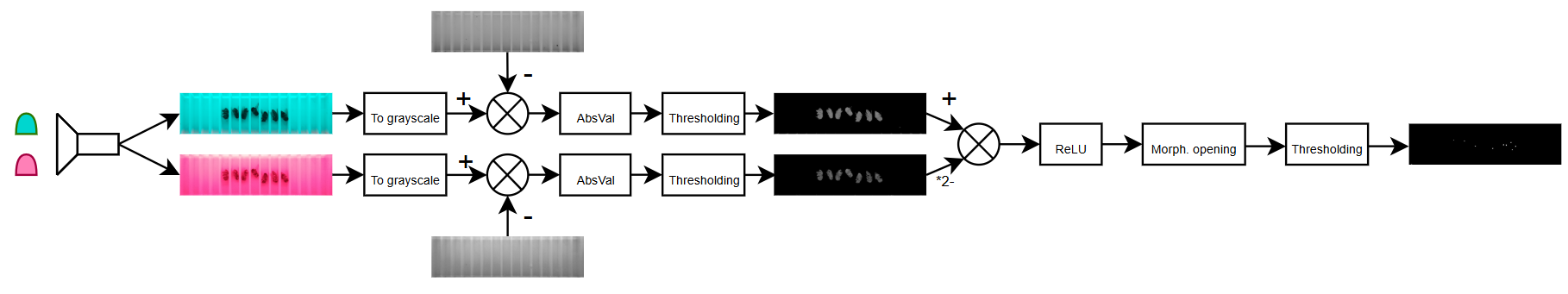}
    \caption{Algorithm of conventional computer vision based approach.}
    \label{fig:CVAlgorithm}
\end{figure*}

After capturing the photos, we convert them into grayscale format and we subtract a background static image from the captured photo. Afterwards, we make an absolute value of these intermediate results and we perform binary thresholding. In the next step, we multiply the infrared image by two, and we subtract it from the previously modified image. Finally, we perform a morphological opening and binary thresholding again. Pixels with the True value belong to Varroa mites and those with the False value belong to the background or bees. The proposed algorithm is shown in the Figure \ref{fig:CVAlgorithm}. Sadly, this approach did not perform as expected, because it is quite hard to define the right thresholds, which are different in every image. Besides it, the performance of our proposed model could be decreased by dirty from pollen or some another insect which can go through tunnels to beehive, for example ants. This method have also a lot of false positive and false negative cases, so we utilized only the results from semantic segmentation, which is described in the section bellow. We assume, that machine learning algorithm will be better, because our proposed conventional method works only with individual pixels, while machine learning method see every pixel value in some context.

\subsection{Semantic segmentation based approach}

In order to find the \textit{Varroa destructor} mites, we utilized a modified U-net semantic segmentation framework available from~\cite{UNetImplementation}. For training, a Cross entropy loss is used.  The network was trained on a computer with CPU AMD Ryzen 5 5600X with 64 GB RAM and GPU NVIDIA GeForce GTX 1060 with 3GB RAM. We trained 40 epochs in total, with batch size=1, learning rate=0.0001. The size of the images was scaled to 0.5 in both dimensions. 10\% of the images were used as validation data. One epoch training took about 2:20 - 2:40 minutes. Because the Varroa mite is not visible in the turquoise images, we decided to train the model only on images of bees illuminated by infrared color.

\section{Experimental results}

\begin{figure}[!h]
    \centering
    \includegraphics[width = 0.5\textwidth]{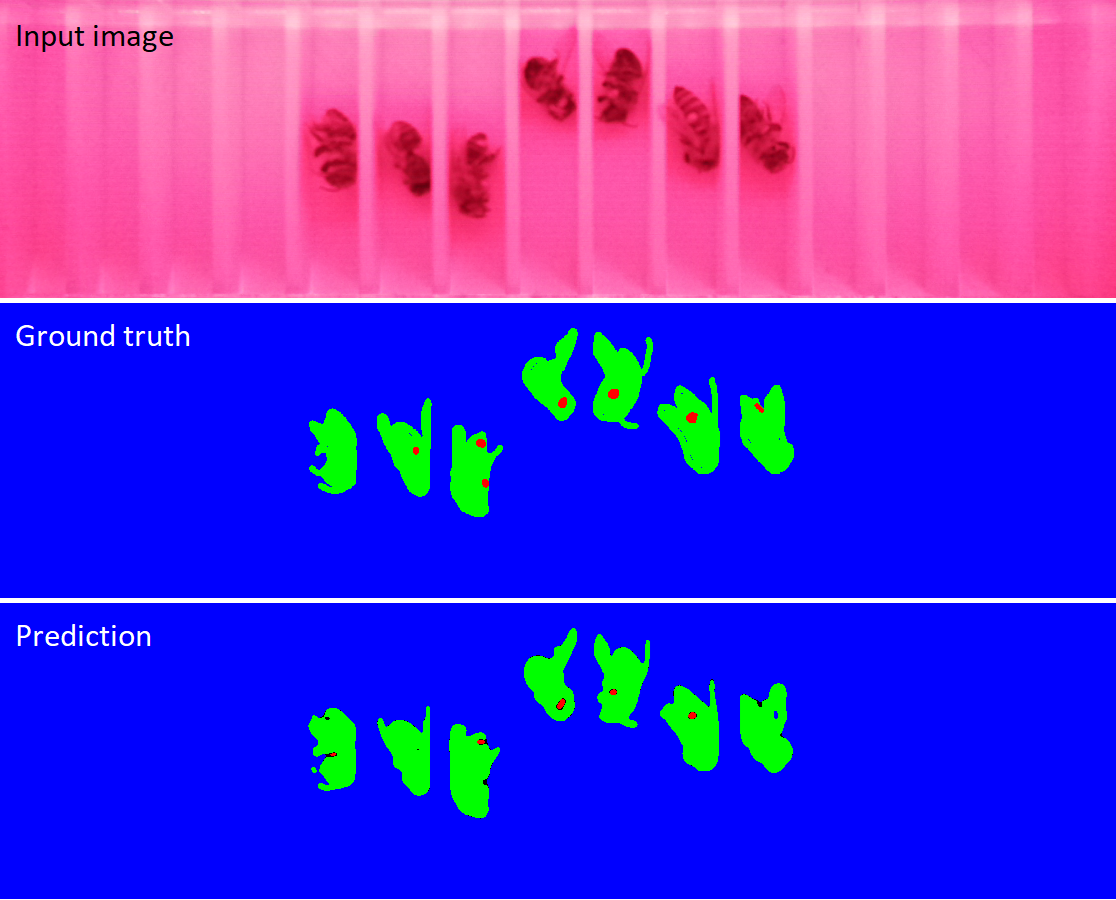}
    \caption{Output of the U-net model.}
    \label{fig:ModelOutputExample}
\end{figure}

In Figure \ref{fig:ModelOutputExample} is shown the output of our model trained on infrared data (780 nm). Our model was trained by 40 epochs. The results were achieved during the epoch 20. Table \ref{tab:ConfMatWithVarroa} left shows a confusion matrix for this model on every image of training data. Table \ref{tab:ConfMatWithoutVarroa} right shows confusion matrix on every training image without the individual mites, because in real case scenario, mites will occur on the bees.

\begin{table}[!h]
    \caption{Confusion Matrix of U-net with (left) and without (right) individual Varroa mites}
    \begin{center}
        \small
        \begin{tabular}{|c|c|c|c|}
            \hline
             -& \multicolumn{2}{c|}{Predicted}\\
             \hline
             \multirow{2}{*}{GT} &TP=1893&FN=588\\
             \cline{2-3}
             &FP=207&TN=0\\
             \hline
    \end{tabular}
    \label{tab:ConfMatWithoutVarroa}
    \quad
    \begin{tabular}{|c|c|c|c|}
            \hline
             -& \multicolumn{2}{c|}{Predicted}\\
             \hline
             \multirow{2}{*}{GT} &TP=954&FN=513\\
             \cline{2-3}
             &FP=204&TN=0\\
             \hline
    \end{tabular}
    \label{tab:ConfMatWithVarroa}
    \end{center}
\end{table}

\section{Discussion}

The goal of our research is to detect the \textit{Varroa destructor} mites as much accurate as possible. On the other hand, the false positive detections should be kept low as the false alarms tasks the beekeeper and they might result in an unnecessary interventions to the hive. Based on this assumption, we record the detected object as Varroa mite only in the case that it contains more pixels than a certain threshold. As greater the area of mite is, the bigger is the confidence of our model. Such filtered results are shown in Table \ref{tab:ConfMatWithoutSmallVarroa}. We can see that the number of false positive cases rapidly decreases up to about 5\% of the original value, but we also increased the number of false negative cases. The probability, that our model will detect real mite is around 55\% as shown in Equation \ref{eq:Prob}, which has to be improved in our future research.


\begin{equation}
\frac{TP}{TP+FN}=\frac{806}{806+661}=55\%
\label{eq:Prob}
\end{equation}

\begin{table}[!h]
    \caption{Confusion Matrix of U-net without mites smaller than 20 pixels}
    \begin{center}
        \small
        \begin{tabular}{|c|c|c|c|}
            \hline
             -& \multicolumn{2}{c|}{Predicted}\\
             \hline
             \multirow{2}{*}{GT} &TP=806&FN=661\\
             \cline{2-3}
             &FP=10&TN=0\\
             \hline

    \end{tabular}
    \label{tab:ConfMatWithoutSmallVarroa}
    \end{center}
\end{table}

Our model reaches inference time to process one capture from 5 up to 6 seconds when using the rPi 4B with Bullseye 64-bit OS and 4GB of RAM. This should be improved by using a more powerful computational hardware, or a rPi in combination with the HW accelerator.

\section{Conclusion}

In this paper, we modified an existing beehive monitoring device to capture images of the bees illuminated by turquoise, infrared, and cold white color. Afterwards, we collected a dataset of photos of dead bees and mites. We utilized these photos to train an U-net, a semantic segmentation architecture. As a result, we reached the probability 55\% that our model will assign real mite pixels to the \textit{Varroa destructor} class. We also tried a conventional computer vision based approach, but the results were not satisfactory and the real use would be limited as our proposed approach requires two overlaying captures of bees under various illumination conditions. Such overlay would not be possible in the real conditions as the bees will move and besides the reasons described in Section~\ref{Se:ConvApp}, it would increase the inaccuracy of this approach.

In our future research, we would like to accelerate our model and improve its accuracy. The high false negative rate is the most problematic aspect of our study, but in our future research, we will try to combine more illuminations in order to better distinguish mites from the background and bees. In combination with an improved loss function, these factors could improve the accuracy and reduce the false negative cases. 

For a practical application, a metric for planning the medical intervention depending on the number of detected mites has to be tested or developed. Such metric has to be developed in cooperation with a veterinary expert, but it could be based for example on the metric presented by~\cite{articleBjerge} in a similar experimental setup.

We also would like to collect a dataset on live bees with a new Raspberry Pi Camera 3 with autofocus, which would solve the blury outputs while capturing in the IR band.


\bibliographystyle{unsrturl}

\vspace{12pt}

\end{document}